\documentclass[10pt,leqno]{article}

\usepackage[utf8]{inputenc} 
\usepackage[T1]{fontenc}    
\usepackage{hyperref}
\usepackage{url}            
\usepackage{booktabs}       
\usepackage{amsfonts}       
\usepackage{nicefrac}       
\usepackage{microtype}      
\usepackage{xcolor}         
\usepackage{graphicx}
\usepackage{amsmath}
\usepackage[ruled,vlined,linesnumbered]{algorithm2e}
\usepackage{tikz}
\usepackage[flushleft]{threeparttable}
\usepackage{amssymb}
\usepackage{wrapfig}
\usepackage{longtable}
\usepackage[numbers]{natbib}
\PassOptionsToPackage{sort&compress}{natbib}
\usepackage[a4paper, total={6in, 8in}]{geometry}

\usetikzlibrary{shapes,decorations,arrows,calc,arrows.meta,fit,positioning}
\tikzset{
    -Latex,auto,node distance =1 cm and 1 cm,semithick,
    state/.style ={ellipse, draw, minimum width = 0.7 cm},
    point/.style = {circle, draw, inner sep=0.04cm,fill,node contents={}},
    bidirected/.style={Latex-Latex,dashed},
    el/.style = {inner sep=2pt, align=left, sloped}
}

\makeatletter
\newcommand\footnoteref[1]{\protected@xdef\@thefnmark{\ref{#1}}\@footnotemark}
\makeatother

\newcommand{\indep}{\!\perp\!\!\!\perp\!}
\newcommand{\notindep}{\not\!\perp\!\!\!\perp\!}
\newcommand{\X}{\ensuremath{\mathbf{X}}}
\newcommand{\given}{\ensuremath{\,|\,}}

\DeclareMathOperator*{\argmin}{arg\,min}

\title{Evaluating LLP Methods: Challenges and Approaches}

\author{%
  Gabriel Franco \\
  Department of Computer Science\\
  Boston University\\
  Boston, MA \\
  \texttt{gvfranco@bu.edu} \\
  \and
  Giovanni Comarela \\
  Department of Informatics \\
  Federal University of Espírito Santo \\
  Vitória, Brazil \\
  \texttt{gc@inf.ufes.br} \\
  \and
  Mark Crovella \\
  Department of Computer Science\\
  Boston University\\
  Boston, MA \\
  \texttt{crovella@bu.edu} \\
}

\begin{document}

\maketitle

\begin{abstract}


Learning from Label Proportions (LLP) is an established machine learning problem with numerous real-world applications. In this setting, data items are grouped into bags, and the goal is to learn individual item labels, knowing only the features of the data and the proportions of labels in each bag.

Although LLP is a well-established problem, it has several unusual aspects that create challenges for benchmarking learning methods. Fundamental complications arise because of the existence of different LLP variants, i.e., dependence structures that can exist between items, labels, and bags. Accordingly, the first algorithmic challenge is the generation of variant-specific datasets capturing the diversity of dependence structures and bag characteristics. The second methodological challenge is model selection, i.e., hyperparameter tuning; due to the nature of LLP, model selection cannot easily use the standard machine learning paradigm. The final benchmarking challenge consists of properly evaluating LLP solution methods across various LLP variants. We note that there is very little consideration of these issues in prior work, and there are no general solutions for these challenges proposed to date.

To address these challenges, we develop methods capable of generating LLP datasets meeting the requirements of different variants. We use these methods to generate a collection of datasets encompassing the spectrum of LLP problem characteristics, which can be used in future evaluation studies. Additionally, we develop guidelines for benchmarking LLP algorithms, including the model selection and evaluation steps. Finally, we illustrate the new methods and guidelines by performing an extensive benchmark of a set of well-known LLP algorithms. We show that choosing the best algorithm depends critically on the LLP variant and model selection method, demonstrating the need for our proposed approach.


\end{abstract}

\vspace{2cm}
\section{Introduction}


In the Learning from Label Proportions (LLP) problem, the goal is to infer a classifier $f: \mathcal{X} \mapsto \mathcal{Y}$ that maps items $\X \in \mathcal{X}$ to labels $Y \in \mathcal{Y}$.  LLP differs from a standard classification problem in that items do not come with individual labels. Instead, during the training phase, the items are provided in groups called `bags', and only the proportions of labels in each bag is available.

Many algorithms have been proposed for LLP, and solutions to the LLP problem have numerous practical applications.  For example, algorithms for LLP have been used to generate fine-grained predictions of public opinion \cite{comarela2018assessing}, to aid in embryo selection during assisted reproduction \cite{hernandez2018fitting}, as a tool in industrial quality control \cite{10.5555/2034161.2034185}, and to infer the demographics of Twitter users \cite{ardehaly2017co}.  

Despite its practical importance, to date standard \emph{methods} and \emph{benchmarks} have not been developed for evaluating LLP algorithms.  We argue that the reasons behind this deficiency stem from the nature of the LLP problem itself.  
As we show in this paper, a meaningful benchmark for LLP must take into consideration a set of issues that do not arise in traditional classification problems.   
Prior work has generally overlooked these issues (as we discuss in \S~\ref{sec:relwork}) and we show in this paper that the relative performance of LLP algorithms generally depends on how these issues are addressed in the evaluation process.  That is, without proper consideration of the issues we raise here, it is possible to draw incorrect conclusions about which LLP algorithm performs best in general.


\begin{figure}[ht]
\centering
\includegraphics[scale=0.75]{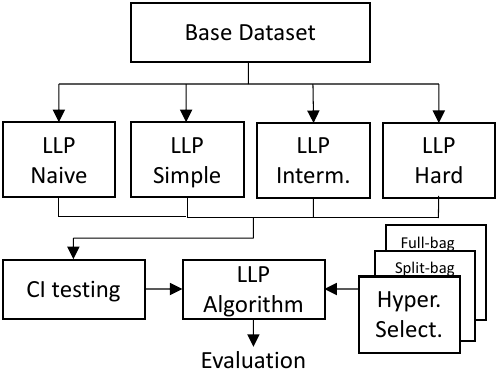}
\caption{Standardized LLP Evaluation.}
\label{fig:diagram-contributions}
\end{figure}

The first issue arises because LLP problems comprise a number of distinct \emph{variants} \cite{kddpaper2023}.  A variant is a particular set of conditional dependence relationships between items, bags, and labels.  Since different LLP algorithms show different relative performance on different variants, it is necessary to benchmark LLP algorithms across a complete set of variants.   


We adopt the approach of generating a set of variants that differ in dependence structure, while maintaining other important aspects constant.  To do so, we start from a single classification dataset (called a `base dataset'), and generate variants through different strategies for assigning items to bags.  In doing so, we also seek to hold properties such as number of bags, bag sizes, and bag label proportions constant across variants (as much as is possible).  Showing how to solve this problem for each variant is our first contribution.  


A related issue concerns how to determine, for any given dataset, which LLP variant it falls under.  To guarantee that any dataset generated follows the chosen variant, we curate a set of independence and conditional independence tests that can be used to verify the dependence structure of each LLP variant. 

The next issue arises because model selection for LLP algorithms involves more complexity than for standard classification problems.  As shown in \cite{kddpaper2023}, there are multiple ways one may test model generalization for LLP, and the choice of which is best depends on the LLP variant.  As a result, careful model selection needs to be built-in to the evaluation process, and we incorporate that in our proposed benchmarking methods.

All of this paper's contributions when combined together allow us to propose a standardized method for evaluating LLP algorithms, which we illustrate in Figure~\ref{fig:diagram-contributions}.   We are distributing to the community the algorithms, code, and datasets to realize the process shown in the Figure.

Finally, we use our standardized benchmarks to evaluate five well-known LLP algorithms from the literature. Using the methods we present here, we generate a standard set of 72 datasets embodying all four variants and a range of characteristics.  The results demonstrate the importance of using the benchmarking strategy we propose.  In particular, we find that different LLP variants lead to the choice of different LLP algorithms and hyperparameter selection strategies.  Thus we conclude that a comprehensive and accurate picture of the performance of any given LLP algorithm requires the use of the entire process shown in Figure~\ref{fig:diagram-contributions}.

\section{Background}

\subsection{LLP and Its Variants} \label{sec:background-variants}


Following the notation of \cite{kddpaper2023}, we define an instance of the LLP problem over $N$ items taken from feature space $\mathcal{X}$; each item is associated to one of $L \leq N$ disjoint groups, called bags. A specific problem instance is given by a set of pairs $D = \{(\mathbf{x}_i, b_i), i = 1, \dots, N\}$ and a matrix $\mathbf{P} \in [0,1]^{L \times C}$, with $\mathbf{x}_i \in \mathcal{X}$ and $b_i \in \{1, \dots, L\}$. We assume that each item $\mathbf{x}_i$ is associated to a label $y_i \in \{1, \cdots, C\}$, which is unknown. The labels are used to define $\mathbf{P}$, where a row $\mathbf{p_\ell} \in [0, 1]^{C}$ is the vector of proportions in bag $\ell$, i.e., $p_{\ell, c} = \frac{|\{i \given b_i = \ell, y_i = c\}|}{|\{i \given b_i = \ell\}|}$. Then, the goal is to learn a classifier that maps the items from feature space $\mathcal{X}$ to $\{1, \cdots, C\}$. We assume the binary classification case, i.e., $C=2$, in all the experiments performed in this paper. However, all the methods presented here apply to any number of classes $C \geq 2$.

An important aspect of any given LLP problem instance is that it can be categorized as a particular \emph{LLP variant.}  We briefly summarize the concepts behind LLP variants here; for detailed discussion we refer to \cite{kddpaper2023}.

We assume that $\{(\mathbf{x}_i, y_i, b_i),\, i = 1,\dots,N\}$ are i.i.d. observations of a random vector $(\mathbf{X}, Y, B)$ with distribution $P_{\mathbf{X}, Y, B} ( \mathbf{x}, y, b)$. Then an LLP variant is defined by the dependence structure between $\X$, $Y$, and $B$, which is described by the following independence and conditional independence tests: 

\begin{equation*}
     \X \indep Y? \quad \X \indep B ? \quad Y \indep B ? \quad X \indep Y \given B? \quad \X \indep B \given Y ? \quad Y \indep B \given \X ?
\end{equation*}


Dependence structures can be derived from directed graphical models (DGMs) \cite{KollerFriedman:GMbook}. Different DGMs can imply the same dependence structure, so for simplicity here we will focus in one specific DGM per variant, which is the one that will be used in this work to generate the datasets. The LLP variants that we will consider are: \emph{Naive, Simple, Intermediate,} and \emph{Hard,} since they are the variants with most significant implications for learning \cite{kddpaper2023}. Table \ref{tbl:taxonomy} shows the dependence structure and the DGM that we will consider for each of these variants.

\begin{table*}[t]
\centering{\footnotesize{

\caption{Dependency structure and DGM of Naive, Simple, Intermediate, and Hard variants}
\label{tbl:taxonomy}
\begin{tabular}{|p{2.1in}|p{0.6in}|c|}
\hline
\multicolumn{1}{|c|}{\textbf{Dependence Structure}} & \multicolumn{1}{|c|}{\textbf{DGM}} & \textbf{LLP Variant} \\
\hline


\parbox{0.5in}{
\centering
\begin{tabular}{lll}
    $\X \notindep Y$ & $\X \indep B$ & $Y \indep B$\\
    $\X \notindep Y \given B$ & $\X \indep B \given Y $& $Y \indep B \given \X$\\
\end{tabular}
} &
\vphantom{\parbox{0.5in}{\centering \scalebox{0.5}{\begin{tikzpicture}
    \node[state] (x) at (0,0) {$\mathbf{X}$};
    \node[state] (y) at (1, 0){$Y$};
    \node[state] (b) at (0.5, -0.75) {$B$};
    \path (x) edge (y);
\end{tikzpicture}%
}}}
\parbox{0.5in}{\centering
\scalebox{0.5}{\begin{tikzpicture}
    \node[state] (x) at (0,0) {$\mathbf{X}$};
    \node[state] (y) at (1, 0){$Y$};
    \node[state] (b) at (0.5, -0.75) {$B$};
    \path (x) edge (y);
\end{tikzpicture}%
}
}
& Naive \\


\hline
\parbox{0.5in}{
\centering
\begin{tabular}{lll}
$ \mathbf{X} \notindep Y $& $ \mathbf{X} \notindep B $& $ Y \notindep B $\\
$ \mathbf{X} \notindep Y \given  B $ & $ \mathbf{X} \indep B \given  Y $& $ Y \notindep B \given  \mathbf{X} $\\
\end{tabular}
} &
\vphantom{\parbox{0.5in}{\scalebox{0.5}{\begin{tikzpicture}
    \node[state] (x) at (0,0) {$\mathbf{X}$};
    \node[state] (y) at (1, 0){$Y$};
    \node[state] (b) at (0.5, -0.75) {$B$};
    \path (x) edge (y);
\end{tikzpicture}%
}}}
\parbox{0.5in}{\centering
\scalebox{0.5}{\begin{tikzpicture}\node[state] (x) at (0,0) {$\mathbf{X}$};\node[state] (y) at (1, 0){$Y$};\node[state] (b) at (0.5, -0.75) {$B$};
    \path (x) edge (y);
    \path (y) edge (b);
\end{tikzpicture}
}
}
& Simple \\


\hline
\parbox{0.5in}{
\centering
\begin{tabular}{lll}
$ \mathbf{X} \notindep Y $& $ \mathbf{X} \notindep B $& $ Y \notindep B $\\
$ \mathbf{X} \notindep Y \given  B $& $ \mathbf{X} \notindep B \given  Y $& $ Y \indep B \given  \mathbf{X} $\\
\end{tabular}
} &
\vphantom{\parbox{0.5in}{\scalebox{0.5}{\begin{tikzpicture}
    \node[state] (x) at (0,0) {$\mathbf{X}$};
    \node[state] (y) at (1, 0){$Y$};
    \node[state] (b) at (0.5, -0.75) {$B$};
    \path (x) edge (y);
\end{tikzpicture}%
}}}
\parbox{0.5in}{\centering
\scalebox{0.5}{\begin{tikzpicture}\node[state] (x) at (0,0) {$\mathbf{X}$};\node[state] (y) at (1, 0){$Y$};\node[state] (b) at (0.5, -0.75) {$B$};
    \path (x) edge (y);
    \path (x) edge (b);
\end{tikzpicture}%
}
}
& Intermediate \\


\hline
\parbox{0.5in}{
\centering
\begin{tabular}{lll}
$ \mathbf{X} \notindep Y $& $ \mathbf{X} \notindep B $& $ Y \notindep B $\\
$ \mathbf{X} \notindep Y \given  B $& $ \mathbf{X} \notindep B \given  Y $& $ Y \notindep B \given  \mathbf{X} $\\
\end{tabular}
} &
\vphantom{\parbox{0.5in}{\scalebox{0.5}{\begin{tikzpicture}
    \node[state] (x) at (0,0) {$\mathbf{X}$};
    \node[state] (y) at (1, 0){$Y$};
    \node[state] (b) at (0.5, -0.75) {$B$};
    \path (x) edge (y);
\end{tikzpicture}%
}}}
\parbox{0.5in}{\centering
\scalebox{0.5}{\begin{tikzpicture}\node[state] (x) at (0,0) {$\mathbf{X}$};\node[state] (y) at (1, 0){$Y$};\node[state] (b) at (0.5, -0.75) {$B$};
    \path (x) edge (y);
    \path (x) edge (b);
    \path (y) edge (b);
\end{tikzpicture}%
}
}
& Hard  \\
\hline
\end{tabular}
}}
\end{table*}

\subsection{Independence and Conditional Independence Tests}


Given a dataset with known labels, in which items have been grouped into bags, it can be important to test which LLP variant the dataset represents.  We assume that $\X \not \indep Y$, which is necessary for there to be a non-trivial algorithm for classification. Hence, we need to perform five tests to identify the variant of a given dataset. Note that in the LLP setting, there are a variety of types of data. The feature vector $\X$ is typically high-dimensional, $Y$ is a binary label (but can also be categorical, in which case $C > 2$), and $B$ is categorical.

In our case, testing whether $Y \indep B$ is straightforward using a contingency table and chi-square test \cite{cressie1984multinomial}. The four remaining cases all involve the high-dimensional random variable $\X$. 

A conditional independence (CI) test attempts to decide, given three random variables $X$, $Y$, and $Z$, whether $X \indep Y \given Z$.  Testing for conditional independence can be challenging when $X$, $Y$, or $Z$ is high-dimensional. A number of recent efforts have approached this problem by using classification based methods \cite{sen2017model,mukherjee2020ccmi,chalupka2018fast}. 

In our work, we use a classification based method called FCIT \cite{chalupka2018fast}. The intuition behind the method is the following: if $X \not \indep Y \given Z$, the quality of the prediction of $Y$ using $X, Z$ should be better than the quality of the prediction of $Y$ using only $Z$. We choose FCIT because it shows stable results for the types of the data encountered in our LLP datasets, it is fast, and it can perform independence tests as well, i.e., testing whether $\X \indep B$.

\section{Related Work} \label{sec:relwork}

When applying LLP algorithms to real-world problems, datasets often do not include labels for the items \cite{comarela2018assessing,hernandez2018fitting,ardehaly2017co}. However, to evaluate LLP algorithms (which is our focus), it is necessary to create LLP datasets from existing supervised learning datasets in which items are equipped with labels.  The strategy taken is to start with a supervised learning dataset and assign items to bags according to some procedure.

Datasets used to date are usually from standard repositories, as UCI%
\footnote{\url{https://archive.ics.uci.edu/ml/datasets.php}}
or LibSVM%
\footnote{\url{https://www.csie.ntu.edu.tw/~cjlin/libsvmtools/datasets/}} 
\cite{quadrianto2009estimating, rueping2010svm, yu2013proptosvm, yu2014learning, patrini2014almost, qi2016learning, chen2017learning, qi2017adaboost, shi2018learning, zhang2019fast, hernandez2019framework, shi2019learning, chen2020ensemble, scott2020learning, qiu2021active, chai2021learning, saket2022combining}. Also, known image classification datasets  \cite{dulac2019deep, liu2021two, shi2018inverse, baruvcic2021fast, chai2021learning, yu2014learning, qi2017adaboost, yu2014modeling, shi2018learning, liu2019learning, tsai2020learning, shi2020deep, liu2022self, liu2022llp, la2022learning, wang2023llp} and other pre-existing collections of datasets \cite{poyiadzi2018label, poyiadzis2019active, xiao2020new, ardehaly2016domain2, qian2019multi, liu2022llp, nandy2022domain, saket2022combining} are used.

Starting from a classification dataset, previous studies have assigned bags to items in a variety of ways. The most common procedure is to randomly assign items to bag, where bags have equal or almost equal size \cite{quadrianto2009estimating, rueping2010svm, yu2013proptosvm, yu2014learning, qi2016learning, qi2017adaboost, chen2017learning, shi2018inverse, shi2018learning, zhang2019fast, shi2019learning, liu2019learning, qian2019multi, dulac2019deep, chen2020ensemble, tsai2020learning, shi2020deep, xiao2020new, scott2020learning, liu2021two, baruvcic2021fast, qiu2021active, chai2021learning, poyiadzis2019active, yu2014modeling, liu2022self, liu2022llp, la2022learning}. Some bag assignments are done in a way to follow a specific bag proportions \cite{poyiadzi2018label, poyiadzis2019active}.  We note that these simple strategies generally create Naive or Simple LLP variants, although this fact has not been emphasized in prior work. Moreover, \cite{quadrianto2009estimating} makes a conditional independence assumption that corresponds to the Simple LLP variant, implying that that paper's datasets may have this dependence structure.

Some studies use a feature to define bags and then removing the feature from the dataset \cite{patrini2014almost, yu2014learning, chai2021learning, nandy2022domain}, and some use a cluster assignment as bags \cite{patrini2014almost, tsai2020learning, saket2022combining}. A more involved procedure is used by \cite{yu2014learning}, where a feature is used as bags, and items are added to bags by sampling with replacement from the set of items that have a specific feature value.  We note that these procedures can create other LLP variants, although again, this fact is not discussed in prior work.


Likewise, the problem of hyperparameter selection in LLP has only been considered in a small number of studies.  The authors in \cite{hernandez2019framework} proposed a $k$-fold based method that assign bags to folds in a way in which each fold have similar proportions in comparison to the entire dataset. Since this method uses a subset of entire bags to train the model and the rest of bags for validation, we call this method \emph{full bag $k$-fold}.   Moreover, \cite{kddpaper2023} proposed new methods to select hyperparameters in LLP: \emph{split-bag bootstrap}, \emph{split-bag shuffle}, and \emph{split-bag $k$-fold}.  We use all of these options in our proposed standardized evaluation method.

The authors in \cite{brahmbhatt2023llp} proposed a set of binary LLP benchmark datasets called LLP-Bench, which is composed of a mix of datasets created in two different ways: randomly assigning items to bags with equal size, and using subsets of features as bags. They use four metrics to measure the hardness of an LLP problem: standard deviation of label proportion, inter vs intra bag separation ratio, mean bag size, and cumulative bag size distribution. They show that their datasets have diversity with respect to these metrics. The datasets are evaluated using nine different LLP algorithms. 

In contrast to LLP-Bench, the methods presented in this paper have a number of advantages.  First, we do not require the base dataset to be tabular in order to generate benchmark datasets. As a result, our benchmark datasets include both image-based and tabular data. Second, we consider problem variants (different dependence structures) in characterizing an LLP benchmark dataset. In particular, we provide general methods that can be used to generate benchmark datasets for four problem variants from any base classification dataset. The desired dependence structures can be verified using methods of conditional independence testing, and we do so for all the benchmark datasets we generate. Generating benchmark datasets spanning the different variants provides important diversity in the benchmarks, as we show in the experimental results below.  Finally, our work provides guidelines for evaluating LLP methods.  Our guidelines take into consideration important issues, in particular guidance on hyperparameter selection and on ensuring the statistical significance of results.

\section{LLP Dataset Generation} \label{sec:dataset-gen}

Benchmark datasets such as MNIST, CIFAR, and ImageNet have played an enormous role in the advance of classification methods in machine learning.  Using benchmark datasets, it is possible to measure the improvement of methods through time, facilitate the comparison between algorithms, and ease  the process of defining the start-of-art. LLP, as a well-established machine learning problem, should have a set of standard benchmark datasets for evaluation of solution strategies. 

As discussed above, the usual approach to creating an LLP dataset is to start from a dataset used for classification.  However, it is not clear how to create LLP datasets from classification datasets with specified dependence structures between items, labels, and bags, i.e., datasets for different LLP variants. Moreover, in an ideal dataset generation process, it should be possible to define the number of bags, the bag proportions, and the sizes of bags. 


In this section, we present methods to generate LLP datasets for different variants given a base (classification) dataset, a specified number of bags, bag proportions, and sizes of bags.  We assume that each item in the base dataset is used once in the resulting LLP dataset, so a minimal condition for feasibility is that bag sizes sum to the dataset size, and expected bag proportions matches the global proportions.  

\subsection{The Dataset Generation Problem}



The input of the problem is as follows: a standard classification dataset $D_c = \{(x_i, y_i), i = 1,\dots,N\}$, where the item-label pairs are i.i.d. observations of a random vector $(\mathbf{X}, Y)$ with distribution $P_{\mathbf{X}, Y} ( \mathbf{x}, y)$; the number of classes $C$; the number of bags, $L$; the size of each bag, i.e., the vector $\mathbf{s}$; the proportions of labels in each bag, i.e., the matrix $\mathbf{P} \in [0, 1]^{L \times C}$, whose rows are stochastic; and a LLP variant, $\nu$. The goal is to associate each $(x_i, y_i)$ to a bag $b_i$, in order to build a LLP instance, such that the number of bags is $L$, the bag proportions and bag sizes obtained are $\mathbf{s}$ and $\mathbf{P}$, respectively, and the dependence structure between items, labels, and bags is $\nu$.

In other words, the goal of the problem is to obtain a data generation mechanism that allows us to sample from a, possibly unknown, probability distribution $Pr(B \given \X, Y)$ which respects the dependence structure imposed by a given LLP variant and the constraints imposed by the bag sizes and bag proportions. In the next sections, we present a way to obtain this sampling mechanism for each LLP variant presented in Table \ref{tbl:taxonomy}.

\subsubsection{Generating Naive LLP}

Considering Table \ref{tbl:taxonomy}, in the Naive LLP case, we have that $Pr(B \given \X, Y) = Pr(B)$, once $Y\indep B$ and $\X \indep B | Y$. Hence, from the problem's input, $Pr(B)$ can be directly computed from $L$ and the vector of bag sizes, $\mathbf{s}$. With $Pr(B)$, bags can be easily sampled and assigned to item-label pairs.


One should note that the desired sampling mechanism can not directly take into account the input matrix of proportions, $\mathbf{P}$, since that would imply $Y \notindep B$. As as consequence, any attempt to generate a Naive LLP dataset in which the components of $\mathbf{P}$ differ significantly of the global proportions in the classification dataset is not feasible.

\subsubsection{Generating Simple LLP}

Using the fact that $\X \indep B \given  Y$, and by the definition of conditional independence, we have $Pr(B \given \X, Y)  = Pr(B \given Y)$. And by Bayes's Theorem, we can write $Pr(B \given Y)$ as $Pr(Y \given B) Pr(B)/Pr(Y)$.


Since $Pr(Y\given B)$ is specified by the bag proportions ($\mathbf{P}$), $Pr(B)$ can be computed from the bag sizes ($\mathbf{s}$), and $Pr(Y)$ can be computed from the labels, the bag assignment rule $Pr(B \given \X, Y)$ can be obtained without needing to consider the features of items.

\subsubsection{Generating Intermediate LLP}\label{sec:intermediate-llp-gen}

Generating an Intermediate LLP instance is more challenging than generating Naive or Simple instances. We can start by rewriting $Pr(B\given\X, Y) = Pr(B\given\X)$, using the fact that $B\indep Y \given \X$. However, in contrast to the Simple LLP case, $Pr(B\given\X)$ is not easily found, since we need to respect the constraints imposed by the desired $P(Y, B).$   The challenge is amplified because $\X$ is likely high-dimensional, and its features can be both continuous and discrete.


In order to overcome these issues, our first step is to cluster the items of the input dataset. The idea is to partition the items into $Q$ clusters ($1, \dots, Q$) in a way that items of the same cluster share some similarity in the feature space. One could imagine simply using the clusters as bags. However, by doing so, there is no guarantee that the constraints related to bag sizes and proportions would be respected (or reasonably approximated). Instead, we use the clusters to simplify the feature space. 

Denote by $Z$ the random variable representing the cluster of a given item. Then, we can write:

\begin{equation} \label{eq:intermediate-prob-derivation}
\begin{split}
Pr(Y, B) &= \sum_z Pr(Y, B, Z = z) = \sum_z Pr(B\given Z = z)Pr(Y\given Z = z)P(Z = z) \\
         &= \sum_z Pr(B\given Z = z)Pr(Y,Z = z).
\end{split}
\end{equation}
where the first line is valid because we assume the following: if $B\indep Y \given \X$, then $B\indep Y \given Z$\footnote{It is plausible to assume it when $Z$ contains all necessary information of $\X$.}.  Then to generate a problem instance, we need to find $Pr(B\given Z = z)$ that satisfies (\ref{eq:intermediate-prob-derivation}); this will give a rule for assigning items to bags based on their cluster ID.  


To solve this, it helps to rewrite (\ref{eq:intermediate-prob-derivation}) as a matrix equation:
$\mathbf{P}_{Y, B} = \mathbf{P}_{Y,Z}\mathbf{P}_{B|Z},$
where $\mathbf{P}_{Y,B} \in \mathbb{R}^{C \times L}$ represents $Pr(Y,B)$, $\mathbf{P}_{Y,Z}\in\mathbb{R}^{C \times Q}$ represents $Pr(Y, Z = z)$, and $\mathbf{P}_{B|Z} \in \mathbb{R}^{Q \times L}$ represents $Pr(B\given Z = z)$.  Note that $\mathbf{P}_{Y,B}$ and $\mathbf{P}_{Y,Z}$ can be easily obtained from the problem's input and from an initial cluster assignment computed over the data.


To generate an Intermediate problem instance we obtain (or approximate) $\mathbf{P}_{B\given Z}$ by solving $\mathbf{P}_{B\given Z} = \argmin_{\mathbf{A}\in\mathcal{A}}\Vert\mathbf{P}_{Y,B} - \mathbf{P}_{Y,Z}\mathbf{A}\Vert_F$, where $\Vert\cdot\Vert_F$ represents the Frobenius norm and the optimization domain $\mathcal{A}$ is the set of all matrices in $\mathbb{R}^{Q\times L}$ whose rows are stochastic.  A locally optimal solution can be found via projected gradient descent, as described in detail in the Supplemental Material (see Appendix \ref{sec:appendix}).  We then use the result of the optimization problem stated above, i.e., $Pr(B\given Z = z)$, as a proxy for the bag assignment rule $Pr(B\given\X)$.

\subsubsection{Generating Hard LLP}

In the Hard LLP variant (see Table \ref{tbl:taxonomy}), $Pr(\X, Y, B)$ cannot be factored, as items, labels, and bags are all correlated with each other, including conditionally. As a consequence, instead of trying to obtain $Pr(B\given \X, Y)$ directly, we create the full joint distribution $Pr(\X, Y, B)$ as a starting point. To that end, we start again with clustering as used in the previous section. 


Again, let $Z$ be the random variable representing the cluster of a given item. Then, the goal is to find a three-dimensional array $\mathbf{P}_{Z,Y,B} \in \mathbb{R}^{Q \times C \times L}$ that encodes $Pr(Z, Y, B)$. To find a suitable $\mathbf{P}_{Z,Y,B}$, we rely on the fact that the marginals $Pr(Z)$, $Pr(Y)$, $Pr(B)$, $Pr(Z, Y)$, and $Pr(Y, B)$ can be computed from the problem's input and the initial cluster assignment.  Thus, we start with a random initialization for $\mathbf{P}_{Z,Y,B}$ which will almost surely meet the desired dependence structure (namely, that no factorization of the distribution is possible).  We then use Iterative Proportional Fitting (IPF, \cite{bishop2007discrete}) to obtain a joint distribution that respects the constraints imposed by the marginals.  Since IPF consists only of rescaling fibers (rows, columns, etc) of the array, it does not introduce any new dependence structure.\footnote{We used \texttt{ipfn} (\url{https://pypi.org/project/ipfn/}) to apply IPF to a three-dimensional table.}

Finally, from $\mathbf{P}_{Z,Y,B}$, we compute $Pr(B\given Z, Y)$, and as in the Intermediate LLP case, we use this conditional distribution as a proxy for the bag assignment rule $Pr(B\given \X, Y)$.

\subsection{Limitations of the Generation Methods} \label{sec:limitations-gen}


For each LLP variant, there are some limitations of the generation methods. For the Naive and Simple variants, the limitations are due to the dependence structure. In the Naive case, since the bags are independent of both items and labels, one can generate only datasets with proportions close to the global proportion. For the Simple variant, the global proportions still imposes constraints, but these are looser: only the expected proportions needs to match the global proportions. 


In the other hand, the main limitation of Intermediate and Hard generation methods is that they rely on a clustering assignment. For the Intermediate variant, each bag contains a mixture of the clusters. As a result the proportions and bag sizes that can be achieved by the generation process are limited to convex combinations of the clusters' proportions and sizes.  This limitation is reflected in the fact that our solution for Intermediate is only locally optimal. For the Hard variant, this limitation is not severe, since we can also use the labels to generate the bags. However, the process continues to rely on a the choice of clustering assignment.

\section{Artifacts} \label{sec:artifacts}

In this section, we describe the artifacts that are provide in this work. We first describe the standard datasets for LLP that we propose; these were developed using the methods described in \S~\ref{sec:dataset-gen}. Additionally, we describe a set of standardized way of evaluating LLP methods, which can be used as a guide for future LLP evaluation studies.

\subsection{Standard Datasets for LLP}\label{sec:standard-datasets}

A set of standard datasets (also called benchmark datasets) for a machine learning problem is important for comparing different algorithms and defining the state-of-art. Although LLP is a well established problem, it lacks of standard datasets which can be widely used to evaluate LLP methods.

To generate a set of standard dataset for LLP, we start from two classification datasets: Adult \cite{misc_adult_2} and CIFAR-10 \cite{krizhevsky2009learning}, which we will call base datasets. Then, we define two numbers of bags: small, with 5 bags, and large, with 10 bags. For the proportions, we defined three types: only bags close to global proportion, only bags far from global proportion, and a mix of both. Also, we defined two different types of bag sizes: all bags with equal size and all bags with different size. Combining these categories of characteristics of a LLP dataset, we can generate 80 datasets in total. For all datasets generated, we used $k$-means with five clusters as a clustering assignment. 

Due to the discussed limitations with the Intermediate LLP generation process (see \S~\ref{sec:limitations-gen}), for the CIFAR-10 we could only generate bags with proportions close to global proportion. This reduced the number of total datasets to 72. A list of all datasets, including the number of bags, proportions, and bag sizes is presented in Appendix \ref{sec:appendix}. This set of standard datasets cover a high number of different characteristics of the LLP problem, enabling a fair comparison between LLP algorithms and hyperparameter selection methods.

Additionally to the set of standard datasets, we provide code to generate datasets for the Naive, Simple, Intermediate, and Hard variants using any given classification dataset. Moreover, we provide the code to verify the dependence structure of a given dataset.\footnote{\label{footnote}The code to reproduce all the results is available at \url{https://github.com/gaabrielfranco/llp-variants-datasets-benchmarks}}

\subsection{Standardized Evaluation Methods} \label{sec:standard-eval-methods}

Compared to standard classification problems, LLP presents additional challenges for algorithm evaluation. Hence, before going into the experimental setup and the results, we discuss aspects of algorithm evaluation specific to LLP. We first note the extra number of features to consider in a LLP dataset: number of bags, bag sizes, and proportions. As discussed in \S~\ref{sec:standard-datasets}, we took into account these aspects in generating the set of standard datasets. It is important that, when we evaluate LLP methods, we test them in different combinations of these features.

Hyperparameter selection is a crucial step in model selection, which is especially challenging for algorithms with a large number of parameters \cite{raschka:arxiv:2018}. When it comes to hyperparameter selection for LLP models, as pointed out by \cite{kddpaper2023}, large datasets with complex dependence structure need a more careful approach to select hyperparameters than the full-bag $k$-fold (see \S \ref{sec:relwork}). Thus, the characteristics of the dataset will lead the choice of the hyperparameter selection strategy used to benchmark LLP algorithms (which we confirm in \S~\ref{sec:eval}). This our proposed methods are intended to support the best choice of the hyperparameter selection strategy.

In the experimental setup, we are also careful with splitting the data between training and testing sets and evaluation metrics. Regarding the train/test split, after splitting the data, we have to assure that the proportions vector passed to the algorithm is the true proportions of the training data. For this reason, we recompute the proportions using the training data, reflecting more a real LLP case. In terms of evaluation metrics, when the ground truth labels are available, it is straightforward to use some supervised learning score, as accuracy or $F_1$-score. However, it is important to take into consideration the same aspects that are taken for the classification literature when it comes to choose the evaluation metric.
As a final result of all these considerations, we present a meta algorithm to evaluate LLP algorithms as Algorithm \ref{alg:meta-alg-llp}. \footnoteref{footnote} 


\begin{algorithm}[ht]
\caption{Meta algorithm for evaluating LLP algorithms}\label{alg:meta-alg-llp}
\KwData{LLP dataset $D = \{(\mathbf{x}_i, b_i), i = 1,\dots,N\}$, associated ground truth labels $y_i, i = 1,\dots,N$, LLP algorithm $\mathbb{A}$, hyperparameter selection strategy $\mathbb{H}$, space of hyperparameters $\theta$}
\KwResult{Performance on the test set of the best trained model}
Split $D$ randomly between training set $D_{\text{train}}$ and testing set $D_{\text{test}}$\\
Compute the proportions $\mathbf{p}$ using $D_{\text{train}}$ and its associated ground truth labels\\
Use a grid search on $\theta$ to find the best combination of hyperparameter $\theta^*$ for $\mathbb{A}$ using $\mathbb{H}$\\
Let $\mathbb{M}$ be the model retrained using $D_{\text{train}}$, $\mathbb{A}$, and $\theta^*$\\
\Return Performance of $\mathbb{M}$ in $D_{\text{test}}$
\end{algorithm}

\section{Evaluation}
\label{sec:eval}

We now put all of the above methods together as shown in Figure~\ref{fig:diagram-contributions}, and conduct an extensive evaluation of LLP algorithms and hyperparameter selection strategies to illustrate the value of our approach.


We use the 72 standardized datasets as detailed in \S~\ref{sec:standard-datasets} to evaluate a set of previously proposed LLP algorithms: EM/LR \cite{comarela2018assessing}, which is based on Logistic Regression; MM \cite{quadrianto2009estimating}, LMM and AMM \cite{patrini2014almost}, which are based on the kernel mean map; and DLLP \cite{ardehaly2017co}, which is a neural network method.

In terms of hyperparameter selection strategies, we used \emph{full-bag $k$-fold} \cite{hernandez2019framework}, \emph{split-bag bootstrap}, \emph{split-bag $k$-fold}, and \emph{split-bag shuffle} \cite{kddpaper2023}. For the EM/LR, we used $C \in \{10^{-2}, \cdots, 10^3\}$. For MM, LMM, and AMM, we used $\lambda \in \{0, 10^0, 10^{1}, 10^{2}\}$. LMM and AMM have additional parameters $\gamma$ and $\sigma$. For LMM and AMM, we used $\gamma \in \{10^{-2}, 10^{-1}, 10^{0}\}$. For LMM, we used $\sigma \in \{2^{-2}, 2^{-1}, 2^{0}\}$ and for AMM we fixed $\sigma=1$. Finally, for DLLP we used $\alpha \in \{10^{-2}, 10^{-3}, 10^{-4}, 10^{-5}, 10^{-6}\}$ with two hidden layers with 100 nodes each.

For each combination of dataset, algorithm, and hyperparameter selection strategy, we run 30 executions of the meta algorithm presented in \S~\ref{sec:standard-eval-methods} (see Algorithm \ref{alg:meta-alg-llp}), using 75\% of the data for training. We measure performance using the $F_1$-score of the learned model in the test set. 

\begin{figure}[ht]
\centering
\includegraphics{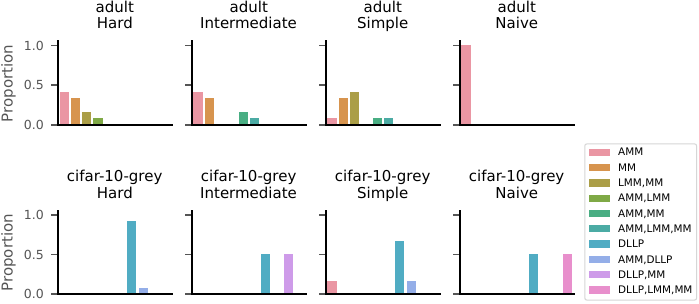}
\caption{Proportion of the best algorithm (set) across datasets and LLP variants.}
\label{fig:best-algorithm-dataset-variant}
\end{figure}

Our experimental setup spans a total of 1,440 experiments, i.e., 360 experiments for each hyperparameter selection strategy.  We trained 1,440$\times$30 = 43,200 models, ie, each algorithm was trained 8,640 times. To handle such large experimental setup, we used the resources from Boston University's Shared Computing Cluster (SCC) \footnote{\url{https://www.bu.edu/tech/support/research/computing-resources/scc/}}.
To compare different algorithms, we consider all executions for the 4 hyperparameter selection strategies (4$\times$30 = 120 executions). Then, we use a $t$-test to determine statistically significant difference in means, starting from the case with highest mean. Figure \ref{fig:best-algorithm-dataset-variant} shows the best algorithm defined by this procedure per base dataset and LLP variant.

\begin{figure}[ht]
\centering
\includegraphics{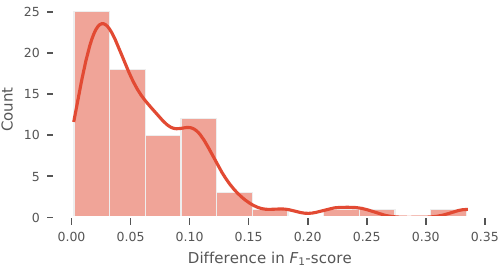}
\caption{Distribution of the $F_1$-score differences between the worst algorithm in the set of the best algorithms and the best algorithm outside the set of the best algorithms.}
\label{fig:effect-sizes}
\end{figure}

Figure~\ref{fig:best-algorithm-dataset-variant} shows that the superiority of various algorithms differs depending on the LLP variant and base dataset.   That is, there is no best algorithm across all variants and base datasets. For instance, we can see that the neural network DLLP is superior for the CIFAR-10 dataset, and its performance is dominant for Hard LLP.  On the other hand, for the adult dataset, mean map methods dominate;  but among those, AMM is superior in the Naive variant, while there is more variety among the best algorithms for the other variants.  
Additionally, Figure~\ref{fig:effect-sizes} shows that the statistically significant differences in performance found between LLP algorithms are typically substantial.

These results show the value of the approach we propose: since the underlying dataset (CIFAR-10 or adult) is constant across variants, as are the bag sizes, label proportions, etc., the differences in performance are clearly due to the differences in LLP variant.  Thus, evaluating across different LLP variants is essential.

\begin{figure}[ht]
\centering
\includegraphics{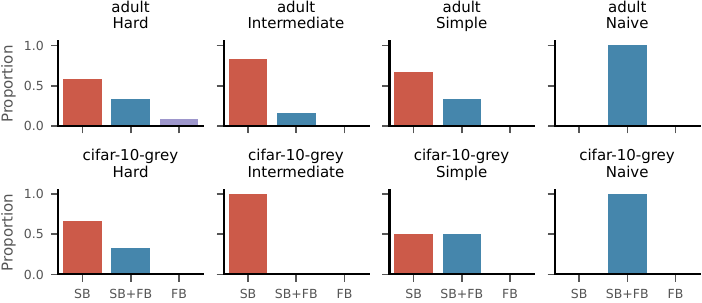}
\caption{Proportion of the best hyperparameter selection method(s) across datasets and LLP variants. SB denotes \textit{split-bag} methods, FB refers to the \textit{full-bag} strategy, while SB+FB means both \textit{split-bag} and \textit{full-bag} strategies.}
\label{fig:best-hyperparam-dataset-variant}
\end{figure}

As a further illustration, Figure~\ref{fig:best-hyperparam-dataset-variant} shows the relative superiority of different hyperparameter selection strategies.  Here too we see that different hyperparameter selection strategies are appropriate for different base datasets and LLP variants (confirming results in \cite{kddpaper2023}), validating the need for considering a range of model selection strategies when evaluating LLP algorithms.

\section{Conclusions}

In conclusion, we argue for the importance of standardizing dataset generation for LLP in a way that acknowledges the dependence structures of LLP variants. Our first contribution is to develop methods to generate LLP variants (Naive, Simple, Intermediate, and Hard) from any given base dataset. We use our methods to generate a set of 72 standard datasets for LLP, which can be used to evaluate future LLP algorithms. Our next contribution is a standardized evaluation method for LLP, which can lead to a fair comparison between different LLP algorithms.  Together, these contributions ease the development of state-of-the-art algorithms for LLP. Finally, we show that LLP methods can have completely different performance depending on the variant or the base classification dataset. These results underscore the importance of evaluating LLP methods using a diverse range of variants and dataset characteristics.


\bibliographystyle{plain}
\bibliography{references}

\newpage

\appendix

\section{Appendix} \label{sec:appendix}

\subsection{Datasets Preparation and Processing}

This paper relies on publicly-available `base' classification datasets to generate LLP benchmark datasets. Detailed documentation on the Adult base dataset is at \cite{misc_adult_2} \footnote{\url{https://archive.ics.uci.edu/dataset/2/adult}} and on the CIFAR-10 based dataset is at \cite{krizhevsky2009learning} \footnote{\url{https://www.cs.toronto.edu/~kriz/cifar.html}}.

Next, we describe the pre-processing steps applied to each base dataset. Following that, we provide a detailed description of how the base datasets were used for LLP benchmark dataset generation.

\subsubsection{Base Datasets}

\paragraph{Adult}   We concatenated the training and testing sets, using all features except \textit{educational-num}. We then one-hot encoded the following features: \textit{age}, \textit{workclass}, \textit{education}, \textit{marital-status}, \textit{occupation}, \textit{relationship}, \textit{race}, and \textit{native-country}. Finally, following \cite{yu2013proptosvm}, we scaled all the features to $[-1, 1]$. This procedure resulted in a dataset with 48842 items and 179 features. 

\paragraph{CIFAR-10 Grey} We concatenated the training and testing sets, and reduced dimensionality by converting images to grayscale using OpenCV \footnote{\url{https://github.com/opencv/opencv-python}}. Then, we flattened the images and merged image groups to form a binary classification problem, classifying vehicles (\textit{airplane}, \textit{automobile}, \textit{ship}, and \textit{truck}) vs animals (\textit{bird}, \textit{cat}, \textit{deer}, \textit{dog}, \textit{frog}, and \textit{horse}). Finally, following \cite{yu2013proptosvm}, we scaled all the features to $[-1, 1]$. The dataset has 60000 items and 1024 features.

\subsubsection{LLP datasets}

The methods and processes for generating LLP benchmark datasets are presented at \S~\ref{sec:dataset-gen} and \S~\ref{sec:standard-datasets}, respectively. Following, we will provide more details of the LLP benchmark dataset generation. 

\paragraph{Adult LLP} 

\begin{itemize}
    \item \textbf{Generation process parameters}
    \begin{itemize}
        \item \textbf{Number of bags}: \textit{small} (5 bags), and \textit{large} (10 bags).
        \item \textbf{Bag sizes}: \textit{equal} (bags with similar size), and \textit{not-equal} (bags with different size).
        \item \textbf{Proportions}: \textit{close-global} (bags with proportions close to the global proportion), \textit{far-global} (bags with proportions far from the global proportion), and \textit{mixed} (mix of bags with proportions close and far from the global proportion). As discussed in \S~\ref{sec:standard-datasets}, the Naive variant does not allow the choice of proportions.
        \item \textbf{Cluster Algorithm}: $k$-means \footnote{\label{footnote-kmeans}\url{https://scikit-learn.org/stable/modules/generated/sklearn.cluster.KMeans.html}} with five clusters (for Intermediate and Hard variants only).
        \item \textbf{Dataset Characteristics}: See Table \ref{tab:llp-datasets-info}.
    \end{itemize}
    \item \textbf{Number of generated datasets}: 40.
    \item \textbf{Format}: Apache Parquet \footnote{\label{footnote-parquet}\url{https://parquet.apache.org/}}.
    \item \textbf{License}: Creative Commons Zero v1.0 Universal.
    \item \textbf{Link to download}: \url{https://github.com/gaabrielfranco/llp-variants-datasets-benchmarks/tree/main/datasets-ci}.
\end{itemize}

\paragraph{CIFAR-10 Grey LLP}

\begin{itemize}
    \item \textbf{Generation process parameters}:
    \begin{itemize}
        \item \textbf{Number of bags}: \textit{small} (5 bags), and \textit{large} (10 bags).
        \item \textbf{Bag sizes}: \textit{equal} (bags with similar size), and \textit{not-equal} (bags with different size).
        \item \textbf{Proportions}: \textit{close-global} (bags with proportions close to the global proportion), \textit{far-global} (bags with proportions far from the global proportion), and \textit{mixed} (mix of bags with proportions close and far from the global proportion). As discussed in \S~\ref{sec:standard-datasets}, the Naive variant does not allow the choice of proportions. Moreover, for this base dataset only, for reasons discussed in \S~\ref{sec:limitations-gen}) it was not possible to generate Intermediate variants with proportions far from \textit{close-global}.  
        \item \textbf{Cluster Algorithm}: $k$-means \footnoteref{footnote-kmeans} with five clusters (for Intermediate and Hard variants only).
        \item \textbf{Datasets Characteristics}: See Table \ref{tab:llp-datasets-info}.
    \end{itemize}

    \item \textbf{Number of generated datasets}: 32.
    \item \textbf{Format}: Apache Parquet \footnoteref{footnote-parquet}.
    \item \textbf{License}: Creative Commons Zero v1.0 Universal.
    \item \textbf{Link to download}: \url{https://github.com/gaabrielfranco/llp-variants-datasets-benchmarks/tree/main/datasets-ci}.
\end{itemize}

The datasets and the code to reproduce the generation process and all the experiments of this paper are available at \url{https://github.com/gaabrielfranco/llp-variants-datasets-benchmarks}. Refer to the README \footnote{\url{https://github.com/gaabrielfranco/llp-variants-datasets-benchmarks/blob/main/README.md}} for instructions ensuring reproducibility. 

\subsection{Algorithms}

As discussed in \S~\ref{sec:intermediate-llp-gen}, we solve the Intermediate LLP generation problem using Projected Gradient Descent (PGD), which is presented in Algorithm \ref{alg:intermediate-llp}.


\begin{algorithm}[ht]
\caption{Projected gradient descent to generate datasets for the LLP intermediate variant}\label{alg:intermediate-llp}
\KwData{Number of bags $L_T$, number of classes $C$, number of clusters $Q$, $\mathbf{P_{yx}} \in \mathbb{R}^{C \times Q}$, $\mathbf{P_{yb}} \in \mathbb{R}^{C \times L_T}$, learning rate $\alpha$, number of maximum iterations $\text{max\_iter}$}
\KwResult{Matrix of probabilities distributions $\mathbf{A} \in \mathbb{R}^{Q \times L_T}$}
$\mathbf{A} = RandomUniformMatrix(Q, L_T)$\\
$\mathbf{A_{old}} = MatrixFull(Q, L_T, 10^5)$\\
$it = 0, tol = 10^{-5}$\\
\While{it $\leq \text{max\_iter}$}{
    $\mathbf{A} = \mathbf{A} - \alpha \left( 2 \mathbf{P_{yx}}^T \mathbf{P_{yx}} \mathbf{A} - 2 \mathbf{P_{yx}}^T \mathbf{P_{yb}} \right)$ \CommentSty{\#Gradient Descent Step}\\
    \CommentSty{\#Projection Step}\\
    $a_{ij} = min(a_{ij}, 1), \forall a_{ij} \in \mathbf{A}$\\
    $a_{ij} = max(a_{ij}, 0), \forall a_{ij} \in \mathbf{A}$\\
    $a_{ij} = \frac{a_{ij}}{\sum_k a_{ik}} , \forall a_{ij} \in \mathbf{A}$\\
    $it = it + 1$\\
    \eIf{$\frac{||\mathbf{A} - \mathbf{A_{old}}||}{|| \mathbf{A} ||} \leq$ tol}{
        break
    }{
        $\mathbf{A_{old}} = \mathbf{A}$\\
    }
}
\Return $\mathbf{A}$
\end{algorithm}

\subsection{Additional results}

We first present the the number of bags, bags sizes, and proportions of the 72 LLP datasets in Table \ref{tab:llp-datasets-info}. Moreover, Table \ref{tab:ci-tests} presents the $p$-value obtained from the CI tests for every LLP dataset. Finally, Table \ref{tab:all-results} presents the $F$-score with 95\% confidence interval for all experiments performed in the paper.

{
\small
\begin{longtable}{p{4cm}p{1cm}p{5cm}p{5cm}}
\caption[Characteristics of LLP Test Cases]{Characteristics of LLP Test Cases\label{tab:llp-datasets-info}}\\

\toprule
                                                                                Dataset & Number of bags &                                                  Proportions &                                                    Bag sizes \\
\midrule
\endfirsthead

\toprule
                                                                                Dataset & Number of bags &                                                  Proportions &                                                    Bag sizes \\
\midrule
\endhead
\midrule
\multicolumn{4}{r}{{Continued on next page}} \\
\midrule
\endfoot

\bottomrule
\endlastfoot
                                   adult-hard-large-equal-close-global-cluster-kmeans-5 &             10 &  (0.34, 0.14, 0.32, 0.15, 0.32, 0.15, 0.31, 0.18, 0.3, 0.19) & (5075, 5251, 5015, 5105, 4809, 4694, 4765, 4912, 4673, 4543) \\
                                     adult-hard-large-equal-far-global-cluster-kmeans-5 &             10 & (0.15, 0.33, 0.09, 0.38, 0.05, 0.44, 0.12, 0.37, 0.07, 0.42) & (5163, 5166, 5136, 4995, 4863, 4621, 4854, 4904, 4638, 4502) \\
                                          adult-hard-large-equal-mixed-cluster-kmeans-5 &             10 &   (0.2, 0.28, 0.14, 0.33, 0.1, 0.39, 0.24, 0.24, 0.05, 0.44) & (5135, 5192, 5098, 4996, 4880, 4666, 4789, 4934, 4665, 4487) \\
                               adult-hard-large-not-equal-close-global-cluster-kmeans-5 &             10 &  (0.34, 0.14, 0.32, 0.15, 0.32, 0.16, 0.3, 0.18, 0.29, 0.21) & (6097, 6241, 5405, 5455, 4832, 4848, 4294, 4378, 3642, 3650) \\
                                 adult-hard-large-not-equal-far-global-cluster-kmeans-5 &             10 & (0.14, 0.34, 0.09, 0.38, 0.04, 0.44, 0.11, 0.37, 0.07, 0.42) & (6217, 6155, 5420, 5436, 4893, 4788, 4328, 4352, 3651, 3602) \\
                                      adult-hard-large-not-equal-mixed-cluster-kmeans-5 &             10 &  (0.19, 0.28, 0.14, 0.34, 0.1, 0.39, 0.24, 0.24, 0.05, 0.44) & (6186, 6141, 5445, 5456, 4908, 4750, 4330, 4376, 3654, 3596) \\
                                   adult-hard-small-equal-close-global-cluster-kmeans-5 &              5 &                               (0.34, 0.13, 0.31, 0.17, 0.24) &                             (10330, 10306, 9519, 9704, 8983) \\
                                     adult-hard-small-equal-far-global-cluster-kmeans-5 &              5 &                                  (0.15, 0.34, 0.1, 0.4, 0.2) &                             (10434, 10210, 9604, 9641, 8953) \\
\end{longtable}

}

{\small
\begin{longtable}{p{4cm}p{1.75cm}p{1cm}p{1.4cm}p{1.4cm}p{1.5cm}p{1.5cm}p{1.5cm}}
\caption[CI tests results for every dataset]{CI tests results for every dataset\label{tab:ci-tests}}\\
\toprule
                                                                                Dataset &  LLP Variant &  Follow DGM & B Indep. Y (p-value) & X Indep. B (p-value) & X Indep. Y Given B (p-value) & X Indep. B Given Y (p-value) & B Indep. Y Given X (p-value) \\
\midrule
\endfirsthead

\toprule
                                                                                Dataset &  LLP Variant &  Follow DGM & B Indep. Y (p-value) & X Indep. B (p-value) & X Indep. Y Given B (p-value) & X Indep. B Given Y (p-value) & B Indep. Y Given X (p-value) \\
\midrule
\endhead
\midrule
\multicolumn{8}{r}{{Continued on next page}} \\
\midrule
\endfoot

\bottomrule
\endlastfoot
                                   adult-hard-large-equal-close-global-cluster-kmeans-5 &         Hard &        True &                 0.00 &                 0.00 &                         0.00 &                         0.00 &                         0.00 \\
                                     adult-hard-large-equal-far-global-cluster-kmeans-5 &         Hard &        True &                 0.00 &                 0.00 &                         0.00 &                         0.00 &                         0.00 \\
                                          adult-hard-large-equal-mixed-cluster-kmeans-5 &         Hard &        True &                 0.00 &                 0.00 &                         0.00 &                         0.00 &                         0.00 \\
                               adult-hard-large-not-equal-close-global-cluster-kmeans-5 &         Hard &        True &                 0.00 &                 0.00 &                         0.00 &                         0.00 &                         0.00 \\
                                 adult-hard-large-not-equal-far-global-cluster-kmeans-5 &         Hard &        True &                 0.00 &                 0.00 &                         0.00 &                         0.00 &                         0.00 \\
                                      adult-hard-large-not-equal-mixed-cluster-kmeans-5 &         Hard &        True &                 0.00 &                 0.00 &                         0.00 &                         0.00 &                         0.00 \\
                                   adult-hard-small-equal-close-global-cluster-kmeans-5 &         Hard &        True &                 0.00 &                 0.00 &                         0.00 &                         0.00 &                         0.00 \\
\end{longtable}
}


{
\small
\begin{longtable}{p{5cm}p{2cm}p{1.5cm}p{1.5cm}p{1.5cm}p{1.5cm}}
\caption[$F$-score with 95\% confidence interval for all experiments]{$F$-score with 95\% confidence interval for all experiments\label{tab:all-results}}\\
\toprule
                                                                                Dataset & Algorithm & Full-Bag K-fold & Split-Bag K-fold & Split-Bag Shuffle & Split-Bag Bootstrap \\
\midrule
\endfirsthead

\toprule
                                                                                Dataset & Algorithm & Full-Bag K-fold & Split-Bag K-fold & Split-Bag Shuffle & Split-Bag Bootstrap \\
\midrule
\endhead
\midrule
\multicolumn{6}{r}{{Continued on next page}} \\
\midrule
\endfoot

\bottomrule
\endlastfoot
                                   adult-hard-large-equal-close-global-cluster-kmeans-5 &       AMM & 0.4731 (0.0239) &  0.5162 (0.0252) &    0.506 (0.0258) &     0.4948 (0.0226) \\
                                   adult-hard-large-equal-close-global-cluster-kmeans-5 &      DLLP & 0.0032 (0.0051) &  0.0017 (0.0015) &    0.0357 (0.035) &     0.0132 (0.0231) \\
                                   adult-hard-large-equal-close-global-cluster-kmeans-5 &     EM/LR & 0.2395 (0.0136) &  0.2448 (0.0133) &   0.2427 (0.0133) &     0.2444 (0.0126) \\
                                   adult-hard-large-equal-close-global-cluster-kmeans-5 &       LMM & 0.4539 (0.0097) &  0.4797 (0.0053) &   0.4841 (0.0042) &      0.4818 (0.004) \\
                                   adult-hard-large-equal-close-global-cluster-kmeans-5 &        MM &  0.4487 (0.012) &  0.4837 (0.0034) &   0.4839 (0.0033) &     0.4829 (0.0033) \\
                                     adult-hard-large-equal-far-global-cluster-kmeans-5 &       AMM & 0.5077 (0.0061) &  0.5116 (0.0044) &   0.5112 (0.0047) &     0.5083 (0.0048) \\
                                     adult-hard-large-equal-far-global-cluster-kmeans-5 &      DLLP & 0.0004 (0.0008) &   0.0113 (0.016) &   0.0139 (0.0208) &     0.0135 (0.0179) \\
                                     adult-hard-large-equal-far-global-cluster-kmeans-5 &     EM/LR &  0.3037 (0.015) &  0.3083 (0.0126) &   0.3053 (0.0137) &     0.3084 (0.0129) \\
\end{longtable}
}

\end{document}